\documentclass[sigconf]{acmart}

\usepackage{graphicx}
\usepackage{url}
\usepackage{amsmath}
\usepackage{amsthm}
\usepackage{booktabs}
\usepackage{algorithm}
\usepackage{algorithmic}
\usepackage{multirow}
\newcommand{\eg}{\textit{e.g.}}
\newcommand{\ie}{\textit{i.e.}}
\DeclareMathOperator*{\argmin}{arg\,min}
\DeclareMathOperator*{\argmax}{arg\,max}

\newcommand{\svec}{{\mathbf{s}}}
\AtBeginDocument{%
  \providecommand\BibTeX{{%
    \normalfont B\kern-0.5em{\scshape i\kern-0.25em b}\kern-0.8em\TeX}}}

\copyrightyear{2021}
\acmYear{2021}
\setcopyright{acmlicensed}\acmConference[KDD '21]{Proceedings of the 27th ACM SIGKDD Conference on Knowledge Discovery and Data Mining}{August 14--18, 2021}{Virtual Event, Singapore}
\acmBooktitle{Proceedings of the 27th ACM SIGKDD Conference on Knowledge Discovery and Data Mining (KDD '21), August 14--18, 2021, Virtual Event, Singapore}
\acmPrice{15.00}
\acmDOI{10.1145/3447548.3467417}
\acmISBN{978-1-4503-8332-5/21/08}

\begin{document}

\fancyhead{}
\title[Toward Deep Supervised Anomaly Detection: Reinforcement Learning from Partially Labeled Anomaly Data]{Toward Deep Supervised Anomaly Detection: \\Reinforcement Learning from Partially Labeled Anomaly Data}


\author{Guansong Pang}
\authornote{Corresponding author: Guansong Pang}
\affiliation{%
  \department{Australian Institute for Machine Learning}
  \institution{The University of Adelaide}
  \city{Adelaide}
  \country{Australia}
  \postcode{5005}
}
\email{pangguansong@gmail.com}

\author{Anton van den Hengel}
\affiliation{%
  \department{Australian Institute for Machine Learning}
  \institution{The University of Adelaide}
  \city{Adelaide}
  \country{Australia}
  \postcode{5005}
}
\email{anton.vandenhengel@adelaide.edu.au}

\author{Chunhua Shen}
\affiliation{%
  \department{Australian Institute for Machine Learning}
  \institution{The University of Adelaide}
  \city{Adelaide}
  \country{Australia}
  \postcode{5005}
}
\email{chunhua.shen@adelaide.edu.au}

\author{Longbing Cao}
\affiliation{%
  \department{Advanced Analytics Institute}
  \institution{University of Technology Sydney}
  \city{Sydney}
  \country{Australia}
  \postcode{2007}
}
\email{longbing.cao@uts.edu.au}


\begin{abstract}

We consider the problem of anomaly detection with a small set of partially labeled anomaly examples and a large-scale unlabeled dataset. This is a common scenario in many important applications. Existing related methods either exclusively fit the limited anomaly examples that typically do not span the entire set of anomalies, or proceed with unsupervised learning from the unlabeled data. We propose here instead a deep reinforcement learning-based approach that enables an end-to-end optimization of the detection of both labeled and unlabeled anomalies. This approach learns the known abnormality by automatically interacting with an anomaly-biased simulation environment, while continuously extending the learned abnormality to novel classes of anomaly (\ie, unknown anomalies) by actively exploring possible anomalies in the unlabeled data. This is achieved by jointly optimizing the exploitation of the small labeled anomaly data and the exploration of the rare unlabeled anomalies. Extensive experiments on 48 real-world datasets show that our model significantly outperforms five state-of-the-art competing methods.

\end{abstract}



\maketitle

\section{Introduction}

Anomaly detection finds applications in a broad range of critical domains, such as intrusion detection in cybersecurity, early detection of disease in healthcare, and fraud detection in finance. Anomalies often stem from diverse causes, resulting in different types/classes of anomaly with distinctly dissimilar features. For example, different types of network attack can embody entirely dissimilar underlying behaviors. By definition, anomalies also occur rarely, and unpredictably, in a dataset. It is therefore difficult, if not impossible, to obtain labeled training data that covers all possible classes of anomaly. This renders fully supervised methods impractical. Unsupervised approaches have dominated this area for decades for this reason \cite{aggarwal2017outlieranalysis}. In many important applications, however, 
there exist a small set of known instances of important classes of anomaly. Despite of the small size, these labeled anomalies provide valuable prior knowledge, enabling significant accuracy improvements over unsupervised methods \cite{tamersoy2014guilt,pang2018repen,pang2019weakly,pang2019devnet,ruff2020deep}. The challenge then is how to exploit those limited anomaly examples without assuming that they illustrate every class of anomaly.

On the other hand, in most application scenarios there is readily accessible large-scale unlabeled data that may contain diverse anomalies from either the same class as the known anomalies, or novel classes of anomaly (\ie, unknown anomalies). Thus, in addition to the anomaly examples, it is also crucial to leverage those unlabeled data for the detection of both known and unknown anomalies. 

In this work we consider the problem of anomaly detection with partially labeled anomaly data, \ie, large-scale unlabeled data (mostly normal data) and a small set of labeled anomalies that only partially cover the classes of anomaly. Unsupervised anomaly detection approaches \cite{liu2012iforest,pang2018repen,breunig2000lof,zong2018autoencoder} can often detect diverse anomalies because they are not limited by any labeled data, but they can produce many false positives due to the lack of prior knowledge of true anomalies. The most related studies are the semi/weakly-supervised approaches \cite{pang2019devnet,ruff2020deep,pang2019weakly} that utilize the labeled anomalies to build anomaly-informed models, but they exclusively fit the limited anomaly examples, ignoring the supervisory signals from possible anomalies in the unlabeled data. A possible solution to this issue is to use current unsupervised methods to detect some pseudo anomalies from the unlabeled data \cite{pang2018repen}, and then feed these pseudo anomalies and the labeled anomalies to learn more generalized abnormality using the semi/weakly-supervised models \cite{pang2019devnet,pang2019weakly,ruff2020deep}. However, the pseudo labeling can have many false positives,
which may deteriorate the exploitation of the labeled anomaly data; moreover, the labeling and the detection modeling are two decoupled steps, failing to jointly optimize the two steps.

To address the problem, this paper proposes an anomaly detection-oriented deep reinforcement learning (DRL) approach that automatically and interactively fits the given anomaly examples and detects known/unknown anomalies in the unlabeled data simultaneously.
Particularly, a neural network-enabled \textit{anomaly detection agent} is devised to exploit the labeled anomaly data to improve detection accuracy, without limiting the set of anomalies sought to those given anomaly examples. The agent achieves this by automatically interacting with a simulated environment created from the labeled and unlabeled data. Most real-world anomaly detection applications involve no sequential decision process (\eg, tabular data), and thus, cannot provide the interactive environment. To tackle this issue, a novel method is introduced to create an \textit{anomaly-biased simulation environment} to enable the agent to effectively exploit the small set of labeled anomaly instances while being deliberately explore the large-scale unlabeled data for any possible anomalies from novel classes of anomaly. We further define a \textit{combined reward} function leveraging supervisory information from the labeled and unlabeled anomalies to achieve a balanced exploration-exploitation. 

We further instantiate the proposed approach into a model called Deep Q-learning with Partially Labeled ANomalies (\textbf{DPLAN}). In DPLAN, the agent is implemented by an adapted version of the well-known deep Q-network (DQN) \cite{mnih2015dqn} specifically designed for anomaly detection. A novel proximity-dependent observation\footnote{Data observations and data instances are used interchangeably in the paper.} sampling method is devised and incorporated into the simulation environment to efficiently and effectively sample next observation.
Further, a labeled anomaly data-based reward and an unsupervised isolation-based reward are synthesized to drive the joint optimization for detecting both of the known and unknown anomalies. 

In summary, this work makes two major contributions.
\begin{itemize}
    \item We propose to tackle a realistic `supervised' anomaly detection problem with partially labeled anomaly data, having the objective to detect both of known and unknown anomalies.
    \item We introduce a novel DRL approach specifically designed for the problem. The resulting anomaly detection agent can automatically and interactively exploit the limited anomaly examples to learn the known abnormality while being actively explore rare unlabeled anomalies to extend the learned abnormality to unknown abnormalities,
    resulting in a joint optimization of the detection of the known and unknown anomalies. To the best of our knowledge, this is the first work tackling such a joint optimization problem. 
    \item We instantiate the proposed approach into a model called DPLAN and extensively evaluate the model on 48 datasets generated from four real-world datasets to replicate scenarios with different coverage of the known abnormality and anomaly contamination rates. The results show that our model performs significantly better and more stably than five state-of-the-art weakly/un-supervised methods, achieving at least 7\%-12\% improvement in precision-recall rates.
\end{itemize}

\section{Related Work}

\noindent \textbf{Anomaly Detection}. Most conventional approaches \cite{breunig2000lof,liu2012iforest,aggarwal2017outlieranalysis} are unsupervised without requiring any manually labeled data and can detect diverse anomalies, but they are often ineffective when handling high-dimensional and/or intricate data. Further, these approaches are often built on some distance/density-based definition of anomalies. Consequently they can produce high false positives when the anomaly definition is mismatched to the data \cite{aggarwal2017outlieranalysis}. Recently deep learning has been explored to enhance the unsupervised approaches, \eg, by learning autoencoder/GANs (generative adversarial networks)-based reconstruction errors to measure normality \cite{zhou2017autoencoder,zong2018autoencoder,schlegl2017gan}, or learning new feature representations tailored for specific anomaly measures \cite{pang2018repen,ruff2018deepsvdd} (see \cite{pang2021deep} for a detailed survey of this area). The most related studies are the weakly-supervised anomaly detection methods \cite{tamersoy2014guilt,pang2018repen,pang2019devnet,pang2019weakly,ruff2020deep} that leverage some partially labeled anomalies to improve detection accuracy, \eg, by label propagation \cite{tamersoy2014guilt}, or end-to-end feature learning \cite{pang2018repen,pang2019devnet,pang2019weakly,ruff2020deep}. One shared issue among these models is that they can be overwhelmingly dominated by the supervisory signals from the anomaly examples, having the risk of overfitting of the known anomalies.

Our problem is related to PU (positive-unlabeled) learning \cite{elkan2008pul,sansone2018pul}, but they are two fundamentally different problems, because the positive instances (\ie, anomalies) in our problem lie in different manifolds or class structures, whereas PU learning assumes the positive instances share the same manifold/structure. Also, the exploration of unlabeled anomalies is related to active anomaly detection \cite{zha2020meta,abe2006active,siddiqui2018kdd},
but we aim at automatically exploring the unlabeled data without human intervention while the latter assumes the presence of human experts for human feedback-guided iterative anomaly detection. Learning with mismatched class distribution \cite{chen2020semi} is related but tackles a very different problem from ours.

\noindent \textbf{DRL-driven Knowledge Discovery}. DRL has demonstrated human-level capability in several tasks, such as Atari 2600 games \cite{mnih2015dqn}. Motivated by those tremendous success,  DRL-driven real-world knowledge discovery emerges as a popular research area. Some successful application examples are recommender systems \cite{zhao2018drl,zheng2018drl} and automated machine learning \cite{zoph2017nas,dong2018hyperparameter}. A related application to anomaly detection is recently investigated in \cite{oh2019sequential}, in which \textit{inverse reinforcement learning} \cite{ng2000irl} is explored for sequential anomaly detection. Our work is very different from \cite{oh2019sequential} in that (i) they focus on unsupervised settings vs. our `supervised' settings; (ii) a sequential decision process is assumed in \cite{oh2019sequential}, largely limiting its applications, whereas our approach does not have such assumptions; and (iii) they aim at learning an implicit reward function whereas we use predefined reward functions to learn anomaly detection agents. Another related application is \cite{li2020autood} that uses DRL to perform neural architecture search for anomaly detection.

\section{The Proposed Approach}

\subsection{Problem Statement}

Unlike the current anomaly-informed models \cite{pang2018repen,pang2019devnet,ruff2020deep} that focus on the supervision information in the small labeled anomaly set, to learn more generalized models, we aim at learning an anomaly detection function driven by the supervisory signals embedded in both the small anomaly examples and the easily accessible large-scale unlabeled data. Specifically, given a training dataset $\mathcal{D}=\{\mathcal{D}^{a},\mathcal{D}^{u}\}$ (with $\mathcal{D}^{a} \cap \mathcal{D}^{u} = \emptyset$) composed by a small labeled anomaly set $\mathcal{D}^{a}$ and a large-scale unlabeled dataset $\mathcal{D}^{u}$, where $\mathcal{D}^{a}$ is spanned by a set of $k$ known anomaly classes while $\mathcal{D}^{u}$ contains mostly normal data and a few anomalies from known and unknown anomaly classes, our goal is then to learn an anomaly scoring function $\phi: \mathcal{D} \rightarrow \mathbb{R}$ that assigns anomaly scores to data instances so that $\phi(\mathbf{s}_{i}) > \phi(\mathbf{s}_{j})$, where $\mathbf{s}_i,\mathbf{s}_j \in \mathcal{D}$ and $\mathbf{s}_{i}$ can be either a known or an unknown anomaly and $\mathbf{s}_{j}$ is a normal instance. 


Note that $\mathcal{D}$ is presumed to be a generic dataset without sequential decision processes, \eg, multidimensional data, because this is the case in most anomaly detection applications.

\subsection{DRL Tailored for Anomaly Detection}



To jointly optimize the detection of known and unknown anomalies, we introduce an anomaly detection-oriented deep reinforcement learning approach, with its key elements including the agent, environment, action space, and rewards specifically designed for anomaly detection. Our key idea is to devise an anomaly detection agent that can fully exploit the supervisory information in the labeled anomaly data $\mathcal{D}^{a}$ to learn generalized known abnormality while being actively explore possible unlabeled known/unknown anomalies in $\mathcal{D}^{u}$ to continuously refine the learned abnormality.

\subsubsection{Foundation of the Proposed Approach}

\begin{figure}[h!]
  \centering
    \includegraphics[width=0.40\textwidth]{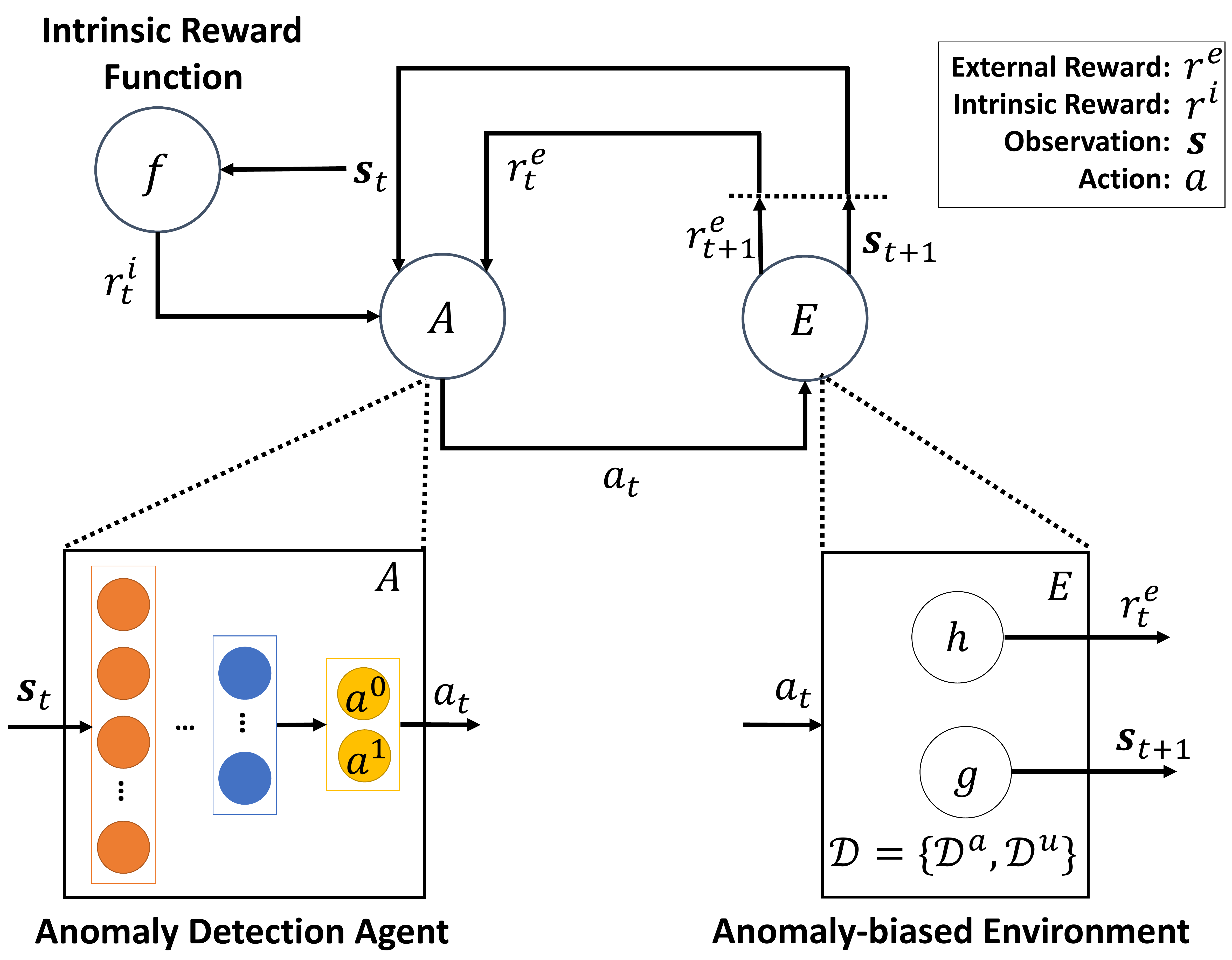}
  \caption{The proposed DRL framework for joint optimization of known and unknown anomaly detection. }
  \label{fig:framework}
\end{figure}

As shown in Figure \ref{fig:framework}, the proposed DRL-based anomaly detection approach consists of three major modules, including an anomaly detection agent $A$, an unsupervised intrinsic reward function $f$, and an anomaly-biased simulation environment $E$ that contains an observation sampling function $g$ and an labeled anomaly-driven external reward function $h$. Our agent $A$ is driven by the combined reward from the $f$ and $h$ functions to automatically interact with the simulation environment $E$ to jointly learn from $\mathcal{D}^{a}$ and $\mathcal{D}^{u}$. To this end, we transform this joint anomaly detection problem into a DRL problem by defining the following key components. 
\begin{itemize}
    \item \textbf{Observation space}. Our observation (or state) space in the environment $E$ is define upon the full training data $\mathcal{D}$, in which each data instance $\mathbf{s}\in \mathcal{D}$ is an observation.
    \item \textbf{Action space}. The action space is defined to be $\{a^{0},a^{1}\}$, with $a^{0}$ and $a^{1}$ respectively corresponding to the action of labeling a given observation $\mathbf{s}$ as `\textit{normal}' and `\textit{anomalous}'.
    \item \textbf{Agent}. The anomaly detection-oriented agent $A$ is implemented by a neural network to seek an optimal action out of the two possible actions $\{a^{0},a^{1}\}$ given an observation $\mathbf{s}$. 
    \item \textbf{Simulation environment}. Since $\mathcal{D}$ is presumed to be generic data, we need to create a simulation environment $E$ to enable meaningful automatic interactions between our agent and the environment. To this end, we define an anomaly-biased \textbf{observation sampling function} $g(\mathbf{s}_{t+1}|\mathbf{s}_t, a_t)$ in the environment, which responds differently to the agent with next observation $\mathbf{s}_{t+1}$ dependent on the observation $\mathbf{s}_{t}$ and the action taken $a_t$ at the time step $t$. The sampling function is designed to bias towards anomalies to better detect known and unknown anomalies.
    \item \textbf{Reward}. We define two reward functions. One is a labeled anomaly data-based \textbf{external reward} $h(r_{t}^{e}|\mathbf{s}_t, a_t)$ that is defined to provide high rewards for the agent when it is able to correctly recognize a labeled anomaly observation, \ie, taking action $a^{1}$ when observing $\mathbf{s}_t \in \mathcal{D}^{a}$. The reward is \textit{external} because it is a pre-defined reward independent of the agent. Further, an unsupervised \textbf{intrinsic reward} function $f(\mathbf{s}_t)$ is defined to measure the novelty of an observation the agent perceives compared to other observations. The agent receives high rewards when it discovers novel observations as a result of self-exploration of the unlabeled data $\mathcal{D}^{u}$. The reward is \textit{intrinsic} in that it is dependent on the self motivation of the agent to explore unexpected observations, a.k.a the agent's curiosity \cite{pathak2017curiosity}. A \textbf{combined external and intrinsic reward} is then defined: $r=c(r^e,r^i)$ to provide an overall reward, where $c$ is a combined function, $r^e$ and $r^i$ are respectively produced by the $h$ and $f$ functions. 
\end{itemize}

By making this transformation, the external reward provides the driving force for the agent to automatically and interactively learn the known abnormality from the labeled anomaly data $\mathcal{D}^{a}$, while the intrinsic reward drives the agent to simultaneously learn unknown abnormalities in the unlabeled data $\mathcal{D}^{u}$.

\subsubsection{Procedure of Our DRL-based Anomaly Detection}

As illustrated in Figure \ref{fig:framework}, our framework works as follows:

\begin{enumerate}
    \item At each time step $t$, 
    the agent $A$ receives an observation $\mathbf{s}_t$ output by the observation sampling function $g$ and takes action $a_t$ to maximize a cumulative reward it may receive. The reward is defined to be proportional to the detection of known/unknown anomalies (see Steps (3)-(4)).
    \item The next observation sampling function $g(\mathbf{s}_{t+1}|\mathbf{s}_t, a_t)$ in the simulation environment $E$ then responds the agent with a new observation $\mathbf{s}_{t+1}$ conditioned on the agent's action on the observation $\mathbf{s}_t$. To effectively leverage both $\mathcal{D}^{u}$ and $\mathcal{D}^{a}$, $g$ is specifically designed to returning possible unlabeled anomalies as many as possible while at the same time equivalently presenting labeled anomaly examples to the agent.
    \item The external reward function $h(r_{t}^{e}|\mathbf{s}_t, a_t)$ further produces a positive reward $r_{t}^{e}$ for the agent if it correctly recognizes the labeled anomaly observation $\mathbf{s}_t$ from $\mathcal{D}^{a}$. This enforces the agent to learn the abnormality of the known anomalies.
    \item The intrinsic reward function $f(\mathbf{s}_t)$ subsequently produces an unsupervised reward $r_{t}^{i}$ for the agent, which encourages the agent to detect novel/surprised observations in the unlabeled data $\mathcal{D}^{u}$ (\eg, unlabeled anomalies).
    \item Lastly, the agent $A$ receives a combined reward $r_t=c(r_{t}^{e},r_{t}^{i})$.
\end{enumerate}

The agent is iteratively trained as in Steps (1)-(5) with a number of episodes, having each \textbf{episode} consisting of a fixed number of observations. To maximize the cumulative combined reward, the agent is optimized to automatically and interactively detect all possible known and unknown anomalies in an unified manner. 

\section{The Instantiated Model: DPLAN}
We instantiate our proposed approach into a model called Deep Q-learning with Partially Labeled ANomalies (DPLAN), with each of its module introduced in detail as follows.

\subsection{DQN-based Anomaly Detection Agent $A$}

Our anomaly detection agent $A$ aims to learn an optimal anomaly detection-oriented action-value function (\ie, Q-value function). Following \cite{mnih2015dqn}, the value function can be approximated as:
\begin{equation}\label{eqn:value}
    Q^{*}(\mathbf{s},a) = \max_{\pi} \mathbb{E}[r_t + \gamma r_{t+1} + \gamma^2 r_{t+2} + \cdots | \mathbf{s}_t=\mathbf{s}, a_{t}=a, \pi], 
\end{equation}
which is the maximum expected return starting from an observation $\mathbf{s}$, taking the action $a \in \{a^{0},a^{1}\}$, and thereafter following a behavior policy $\pi=P(a|\mathbf{s})$, with the \textbf{return} defined as the sum of rewards $r_t$ discounted by a factor $\gamma$ at each time step $t$. Different off-the-shelf DRL algorithms can be used to learn $Q^{*}(\mathbf{s},a)$. In this work, the well-known deep Q-network (DQN) \cite{mnih2015dqn} is used, which uses deep neural networks as the function approximator with the parameters $\theta$: $Q(\mathbf{s},a;\theta)=Q^{*}(\mathbf{s},a)$. It then learns the parameters $\theta$ by iteratively minimizing the following loss:
\begin{equation}\label{eqn:dqn}
    L_j(\theta_j)=\mathbb{E}_{(\mathbf{s},a,r,\mathbf{s}^{\prime})\sim U(\mathcal{E})}\bigg[\Big(r+\gamma\max_{a^{\prime}}Q(\mathbf{s}^{\prime},a^{\prime};\theta_j^{-})-Q(\mathbf{s},a;\theta_j)\Big)\bigg],
\end{equation}
where $\mathcal{E}$ is a set of the agent's learning experience with each element stored as $e_t=(\mathbf{s}_t,a_t,r_t,\mathbf{s}_{t+1})$; the loss is calculated using minibatch samples drawn uniformly at random from the stored experience; $\theta_j$ are the parameters of the Q-network at iteration $j$; the network with the parameters $\theta_j^{-}$ is treated as a target network to compute the target at iteration $j$, having $\theta_j^{-}$ updated with $\theta_j$ every $K$ steps.


\subsection{Proximity-driven Observation Sampling $g$}

The sampling function $g$, a key module in the environment $E$, is composed by two functions, $g_{a}$ and $g_{u}$, to empower a balanced exploitation and exploration of the full data $\mathcal{D}$. Particularly, $g_{a}$ is a function that uniformly samples $\mathbf{s}_{t+1}$ from $\mathcal{D}^{a}$ at random, \ie, $\mathbf{s}_{t+1} \sim U(\mathcal{D}^{a})$, which offers the same chance for each labeled anomaly to be exploited by the agent. 

On the other hand, $g_{u}$ is a function that samples $\mathbf{s}_{t+1}$ from $\mathcal{D}^{u}$ based on the proximity of the current observation. To enable effective and efficient exploration of $\mathcal{D}^{u}$, $g_{u}$ is defined as 
\begin{equation}\label{eqn:observation}
g_{u}(\mathbf{s}_{t+1}|\mathbf{s}_t, a_t;\theta^e) =
  \begin{cases}
    \argmin\limits_{\mathbf{s}\in \mathcal{S}} d\big(\mathbf{s}_t, \mathbf{s};\theta^e\big)       &\text{if } a_t=a^1\\
    \argmax\limits_{\mathbf{s}\in \mathcal{S}} d\big(\mathbf{s}_t, \mathbf{s};\theta^e\big)  &\text{if } a_t=a^0,
  \end{cases}
\end{equation}
where $\mathcal{S} \subset \mathcal{D}^{u}$ is a random subsample, $\theta^e$ are the parameters of $\psi(\cdot;\theta^e)$ that is a feature embedding function derived from the last hidden layer of our DQN, and $d$ returns a Euclidean distance between $\psi(\mathbf{s}_t;\theta^e)$ and $\psi(\mathbf{s};\theta^e)$ to capture the distance perceived by the agent in its representation space.

Particularly, $g_{u}$ returns the nearest neighbor of $\mathbf{s}_t$ when the agent believes the current observation $\mathbf{s}_t$ is an anomaly and takes action $a^1$. This way allows the agent to explore observations that are similar to the suspicious anomaly observations in the labeled data $\mathcal{D}^{u}$. $g_{u}$ returns the farthest neighbor of $\mathbf{s}_t$ when $A$ believes $\mathbf{s}_t$ is a normal observation and takes action $a^0$, in which case the agent explores potential anomaly observations that are far away from the normal observation. Thus, both cases are served for effective active exploration of the possible anomalies in the large $\mathcal{D}^{u}$. 

The parameters $\theta^e$ are a subset of the parameters $\theta$ in DQN. The nearest and farthest neighbors are approximated on subsample $\mathcal{S}$ rather than $\mathcal{D}^{u}$ for efficiency consideration, and we found empirically that the approximation is as effective as performing $g_u$ on the full $\mathcal{D}^{u}$. $|\mathcal{S}|=1000$ is set by default. $\mathcal{S}$ and $\theta^e$ are constantly updated to compute $d$ for each step.

During the agent-environment interaction, both $g_{a}$ and $g_{u}$ are used in our simulator: with probability $p$ the simulator performs $g_{a}$, and with probability $1-p$ the simulator performs $g_{u}$. This way enables the agent to sufficiently exploit the small labeled anomaly data while exploring the large unlabeled data. In this work $p=0.5$ is used to allow both labeled anomalies and unlabeled data to be equivalently harnessed.

\subsection{Combining External and Intrinsic Rewards}

\subsubsection{Labeled Anomaly Data-based External Reward Function $h$}

The below $h$ function is defined to yield a reward signal $r_{t}^{e}$ to our agent based on its performance on detecting known anomalies:
\begin{equation}\label{eqn:externalreward}
r_{t}^{e}=h(\mathbf{s}_t, a_t) =
  \begin{cases}
    1       &\text{if } a_t=a^1 \text{ and } \mathbf{s}_t \in \mathcal{D}^{a} \\
    0       &\text{if } a_t=a^0 \text{ and } \mathbf{s}_t \in \mathcal{D}^{u} \\
    -1      &\text{otherwise}.
  \end{cases}
\end{equation}
It indicates that the agent receives a positive $r^{e}$ only when it correctly labels the known anomalies as `\textit{anomalous}'. The agent receives no reward if it correctly recognize the normal observations, and it is penalized with a negative reward for either false-negative or false-positive detection. Thus, $r^{e}$ explicitly encourages the agent to fully exploit the labeled data $\mathcal{D}^{a}$.

To maximize the return, the agent is driven to interactively learn the known abnormality to achieve high true-positive detection and avoid false negative/positive detection. This learning scheme enables DPLAN (with this external reward alone) to leverage the limited anomaly examples better than the existing semi-supervised methods \cite{pang2019devnet,ruff2020deep}, as shown in our experiments in Section \ref{subsec:ablation}.


\subsubsection{Unsupervised Intrinsic Reward Function $f$}

Unlike $r^e$ that encourages the exploitation of the labeled data $\mathcal{D}^{a}$, the intrinsic reward $r^i$ is devised to encourage the agent to explore possible anomalies in the unlabeled data $\mathcal{D}^{u}$. It is defined as
\begin{equation}\label{eqn:iforest}
  r^i_t=f(\mathbf{s}_t;\theta^e)=\text{iForest}\big(\mathbf{s}_t;\theta^e\big),
\end{equation}
where $f$ measures the abnormality of $\mathbf{s}_t$ using the well-known isolation-based unsupervised anomaly detector, iForest \cite{liu2012iforest}. Isolation is defined by the number of steps required to isolate an observation $\mathbf{s}$ from the observations in $\mathcal{D}^{u}$ through half-space data partition. iForest is used here because it is computationally efficient and excels at identifying rare and heterogeneous anomalies.

Similar to $g_u$ in Eq. (\ref{eqn:observation}), the $f$ function also operates on the low-dimensional $\psi$ embedding space parameterized by $\theta^e$. That means both the training and inference in iForest are performed on the $\psi$-based projected data (\ie, $\psi(\mathcal{D}^{u};\theta^e)$). This enables us to capture the abnormality that is faithful w.r.t. our agent. This also guarantees iForest always works on low-dimensional space as it fails to work effectively in high-dimensional space \cite{liu2012iforest}. The output of iForest is rescaled into the range $[0, 1]$, and we accordingly have $r^i \in [0, 1]$, with larger $r^i$ indicating more abnormal. Thus, regardless of the action taken, our agent receives large $r^i$ whenever the agent believes the observation is rare or novel compared to previously seen observations. This way helps the agent detect possible unlabeled anomalies in $\mathcal{D}^{u}$. To balance the importance of exploration and exploitation, the overall reward the agent receives at each time step $t$ is defined as
\begin{equation}\label{eqn:overallreward}
    r_t=r^e_t + r^i_t.
\end{equation}

\subsection{Theoretical Analysis of DPLAN}

During training, the agent $A$ in DPLAN is trained to minimize the loss in Eq. (\ref{eqn:dqn}) in an end-to-end fashion. Let $Q(\mathbf{s},a;\theta^{*})$ be the Q-network with the learned $\theta^{*}$ after training, then at the inference stage, $Q(\hat{\svec},a;\theta^{*})$ outputs an estimated value of taking action $a^{0}$ or $a^{1}$ given a test observation $\hat{\svec}$. Since $a^{1}$ corresponds to the action of labeling $\hat{\svec}$ as `\textit{anomalous}', $Q(\hat{\svec},a^{1};\theta^{*})$ can be used as anomaly score. The intuition behind this scoring is discussed as follows.

Let $\pi$ be a policy derived from $Q$, then the expected return of taking the action $a^{1}$ given the observation $\hat{\svec}$ under the policy $\pi$, denoted by $q_{\pi}(\hat{\svec}, a^{1})$, can be defined as
\begin{equation}\label{eqn:expectedreturn}
    q_{\pi}(\hat{\svec}, a^{1}) = \mathbb{E}_{\pi}\bigg[\sum_{n=0}^{\infty}\gamma^{n}r_{t+n+1}\Big|\hat{\svec}, a^{1}\bigg].
\end{equation}
Let $\hat{\svec}^{i}$, $\hat{\svec}^{j}$ and $\hat{\svec}^{k}$ be labeled anomaly, unlabeled anomaly and unlabeled normal observations respectively, we have $h(\hat{\svec}^i, a^1)>h(\hat{\svec}^j, a^1)>h(\hat{\svec}^k, a^1)$. $f(\hat{\svec}^i;\theta^e) \approx f(\hat{\svec}^j;\theta^e) > f(\hat{\svec}^i;\theta^e)$ also holds provided that $f$ well captures the abnormality of the three observations. Since $r_t$ in Eq. (\ref{eqn:expectedreturn}) is the sum of the outputs of the $h$ and $f$ functions, $q_{\pi}(\hat{\svec}^{i}, a^{1})>q_{\pi}(\hat{\svec}^{j}, a^{1})>q_{\pi}(\hat{\svec}^{k}, a^{1})$ holds under the same policy $\pi$. Thus, when the agent well approximates the $Q$-value function after a sufficient number of training time steps, its estimated returns yield: $Q(\hat{\svec}^i,a^{1};\theta^{*}) > Q(\hat{\svec}^j,a^{1};\theta^{*}) > Q(\hat{\svec}^k,a^{1};\theta^{*})$; so the observations with large $Q(\hat{\svec},a^{1};\theta^{*})$ are anomalies of our interest. 

\section{Experiments}

\subsection{Datasets}
Unlike prior work \cite{liu2012iforest,ting2017defying,siddiqui2018kdd,pang2018repen,pang2019devnet} where many datasets contain only one anomaly class, our datasets need to contain at least two anomaly classes since we assume there are both known and unknown anomaly classes. As shown in Table \ref{tab:datainfo}, a pool of four widely-used real-world datasets from four diverse domains is used, including UNSW\_NB15 \cite{pang2019devnet} from network intrusion, Annthyroid \cite{liu2012iforest,siddiqui2018kdd,pang2019devnet,ruff2020deep} from disease detection, HAR \cite{ting2017defying} from human activity recognition, and Covertype \cite{liu2012iforest,ting2017defying,siddiqui2018kdd,akoglu2012fast} from forest cover type prediction. After preprocessing,
UNSW\_NB15 contains seven anomaly classes and the other three datasets contains two anomaly classes, with each class be (semantically) real anomalies.

\begin{table}[htbp]
\caption{Statistics of original four datasets. $D$ is the data dimensionality. Each dataset contains 2-7 anomaly classes.}
\scalebox{0.75}{
\begin{tabular}{  l@{}|l@{}|l@{}c|l@{}c }
\hline
\multicolumn{2}{ c}{\textbf{Dataset}}& \multicolumn{2}{c}{\textbf{Normal Class}} & \multicolumn{2}{ c }{\textbf{Anomaly Class}} \\\hline
Data Name & $D$ & Class Name & Class Size  &  Class Name & Class Size (\%) \\ \hline
\multirow{7}{6em}{UNSW\_NB15} & \multirow{7}{2em}{196} & \multirow{7}{6em}{normal network flows} & \multirow{7}{2em}{93,000} & analysis & 2,677 (2.80\%) \\ 
  & &  & & backdoor & 2,329 (2.44\%)    \\ 
  & &  & & DoS &  3,000 (3.13\%)   \\ 
  & &  &   & exploits& 3,000 (3.13\%)    \\ 
  & &  &   & fuzzers & 3,000 (3.13\%)    \\ 
  & &  &   & generic& 3,000 (3.13\%)   \\ 
  & &  &   &reconnaissance & 3,000 (3.13\%)  \\ \hline
\multirow{2}{6em}{Annthyroid} & \multirow{2}{2em}{21} & \multirow{2}{6em}{normal patients}&\multirow{2}{2em}{6,666}& hypothyroid &  166 (2.43\%) \\ 
  & &  &  & subnormal & 368 (5.23\%)   \\ \hline
\multirow{2}{6em}{HAR} & \multirow{2}{2em}{561} & \multirow{2}{8em}{walking, sitting, standing, laying} & \multirow{2}{2em}{7,349} & downstairs & 150 (2.00\%) \\
  & &  & & upstairs & 150 (2.00\%)  \\\hline
\multirow{2}{6em}{Covertype} & \multirow{2}{2em}{54} & \multirow{2}{8em}{the largest class (lodgepole pine)} & \multirow{2}{2em}{283,301} & cottonwood& 2,747 (0.96\%) \\
  & &  &   & douglas-fir & 17,367 (5.78\%)  \\\hline
\end{tabular}
}
\label{tab:datainfo}
\end{table}

These four datasets serve as a base pool of our experiments only. They are leveraged in Section \ref{subsec:realworld} to create 48 datasets to evaluate the anomaly detection performance in different scenarios. 

\subsection{Competing Methods and Their Settings}

DPLAN is compared with five state-of-the-art competing anomaly detectors below:
\begin{itemize}
    \item \textbf{DevNet} \cite{pang2019devnet} is a deep detector that leverages a few labeled anomalies and a Gaussian prior over anomaly scores to perform end-to-end anomaly detection.
    \item \textbf{Deep SAD} \cite{ruff2020deep} is a deep semi-supervised method using a small number of both labeled normal and anomalous instances. Following \cite{tax2004svdd,ruff2020deep}, Deep SAD is adapted to our setting by enforcing a margin between the one-class center and the labeled anomalies while minimizing the center-oriented hypersphere. We found that Deep SAD significantly outperforms its shallow version \cite{gornitz2013semisupervisedad} by over 30\% AUC-PR improvement. Thus, we report the results of Deep SAD only. 
    \item \textbf{REPEN} \cite{pang2018repen} is a recent deep unsupervised detector that learns representations specifically tailored for distance-based anomaly measures. Another popular deep unsupervised detector DAGMM \cite{zong2018autoencoder} is also tested, but it is less effective than REPEN. Thus we focus on REPEN.
    \item \textbf{iForest} \cite{liu2012iforest} is a widely-used unsupervised method that detects anomalies based on how many steps are required to isolate the instances by random half-space partition.
    \item \textbf{DUA} is a variant of DevNet with an additional Unlabeled Anomaly detector, \ie, it is DevNet trained with the labeled anomaly set and pseudo anomalies identified in the unlabeled data. To have a straightforward comparison, DUA uses the same unlabeled anomaly explorer as DPLAN: iForest. iForest returns a ranking of data instances only. A cutoff threshold, \eg, top-ranked $n\%$ instances, is required to obtain the pseudo anomalies. $n=\{0.01, 0.05, 0.1, 0.5, 1, 2, 4\}$ are probed. We report the best results achieved using the threshold $0.05\%$.
\end{itemize}

A multilayer perceptron network is used in the Q-network since the experiments focus on tabular data. All competing deep methods worked effectively using one hidden layer but failed to work using a deeper network due to the limit of the small labeled data. To have a pair comparison, all deep methods use one hidden layer with $l$ units and the ReLU activation \cite{nair2010rectified} by default. Following \cite{pang2018repen,pang2019devnet}, $l=20$ is used. DPLAN can also work effectively with deeper architectures (see our ablation study in Section \ref{subsec:ablation} for detail).

DPLAN is trained with 10 episodes by default, with each episode consisting of 2,000 steps. 10,000 warm-up steps are used. The target network in DQN is updated every $K=10,000$ steps. The other optimization settings of DPLAN are set to the default settings in the original DQN. DevNet (and DUA) and REPEN are used with the settings respectively recommended in \cite{pang2019devnet,pang2018repen}. Deep SAD uses the same optimization settings as DevNet, which enable it to obtain the best performance. The isolation trees with the recommended settings \cite{liu2012iforest} are used in iForest, Eq. (\ref{eqn:iforest}) in DPLAN, and DUA. 

\subsection{Performance Evaluation Measures}
Two widely-used complementary measures, including the Area Under Receiver Operating Characteristic Curve (AUC-ROC) and Area Under Precision-Recall Curve (AUC-PR) \cite{boyd2013aucpr}, are used. AUC-ROC, which summarizes the ROC curve of true positives against false positives, is widely-used due to its good interpretability, but it often presents an overoptimistic view of the detection performance; whereas AUC-PR summarizes the curve of precision and recall, which focuses on the performance on the anomaly class only and is thus much more indicative when the anomalies are of our interest only. 
The reported AUC-ROC and AUC-PR are averaged results over 10 independent runs. The paired \textit{Wilcoxon} signed rank using the AUC-ROC (AUC-PR) across multiple datasets is used to examine the statistical significance of the performance of our method.


\subsection{Known and Unknown Anomaly Detection in Real-world Datasets}\label{subsec:realworld}

\subsubsection{Scenario I: Known Anomalies from One Class}
We split each of the \textit{NB15}, \textit{Thyroid}, \textit{HAR} and \textit{Covertype} datasets into training and test sets, with 80\% data of each class into the training data and the other 20\% data into the test set. For the training data we retain only a few labeled anomalies to be $\mathcal{D}^a$, and randomly sample some anomalies from each anomaly class and mix them with the normal training instances to produce the anomaly-contaminated unlabeled data $\mathcal{D}^u$. We then create 13 datasets, with each dataset having $\mathcal{D}^a$ sampled from only \textit{one specific anomaly class}; these datasets are shown in Table \ref{tab:aucprroc}, where each dataset is named by the known anomaly class. The test data is fixed after the data split, which contains \textit{one known and one-to-six unknown anomaly classes}, accounting for 0.96\%-5.23\% of the test data. Since only a small number of labeled anomalies are available in many applications, in each dataset the number of labeled anomalies is fixed to 60, accounting for 0.03\%-1.07\% of the training data. Anomalies are rare events, so the anomaly contamination rate in $\mathcal{D}^{u}$ is fixed to 2\%. See Section \ref{subsec:increasingknownclasses} (\ref{subsec:contamination}) for varying $|\mathcal{D}^a|$ (contamination rates).

The AUC-PR and AUC-ROC results on the 13 datasets are shown in Table \ref{tab:aucprroc}. DPLAN achieves the best performance on 10 datasets in terms of both AUC-PR and AUC-ROC, with the performance on the other two datasets close to the best performer. Particularly, in terms of AUC-PR, on average, DPLAN substantially outperforms the anomaly-informed detectors DevNet (7\%), Deep SAD (11\%) and DUA (12\%), and obtains nearly 100\% improvement over both of the unsupervised anomaly detectors. In terms of AUC-ROC, DPLAN substantially outperforms all contenders by about 1\%-13\%. The improvement of DPLAN in AUC-PR over all counterparts is significant at the 99\% confidence level; the improvement in AUC-ROC is also significant at least at the 90\% confidence level.

Further, DPLAN performs very stably across all 13 datasets, having significantly smaller AUC standard deviation than DevNet, Deep SAD, and DUA, \ie, averagely 0.004 vs. 0.024/0.027/0.023 in AUC-PR and 0.003 vs. 0.019/0.017/0.010 in AUC-ROC.

Although the labeled anomalies account for only a tiny proportion of the training data, \ie, 0.03\%-1.07\%, DPLAN (as well as DevNet, Deep SAD, and DUA) can leverage the supervision information given by these anomaly examples to substantially enhance the true positives, significantly outperforming the unsupervised REPEN and iForest, especially in AUC-PR. Thus, below we focus on the comparison between DPLAN and the three best contenders.

\begin{table*}[htbp]
\caption{AUC-PR and AUC-ROC performance (mean$\pm$std) of DPLAN and five competing methods on 13 real-world datasets. }
\centering 
\scalebox{0.75}{
\begin{tabular}{ l@{}|p{1.35cm}p{1.35cm}p{1.45cm}p{1.35cm}p{1.35cm}p{1.45cm}|p{1.35cm}p{1.35cm}p{1.45cm}p{1.35cm}p{1.35cm}c@{}}
\hline
&    \multicolumn{6}{ c|}{\textbf{AUC-PR Performance}} & \multicolumn{6}{ c }{\textbf{AUC-ROC Performance}}  \\ \hline
Dataset &  \centering\textbf{DPLAN}  & \centering\textbf{DevNet}  & \centering\textbf{Deep SAD}  & \centering\textbf{REPEN} & \centering\textbf{iForest} & \centering\textbf{DUA}  &  \centering\textbf{DPLAN}  & \centering\textbf{DevNet}  & \centering\textbf{Deep SAD}  & \centering\textbf{REPEN} & \centering\textbf{iForest}& \textbf{DUA}\\\hline
Analysis & \textbf{0.683}$\pm$0.008 & 0.640$\pm$0.033 & 0.595$\pm$0.036 & 0.447$\pm$0.008 & 0.378$\pm$0.022 & 0.609$\pm$0.010 & \textbf{0.852}$\pm$0.004 & 0.839$\pm$0.052 & 0.769$\pm$0.019 & 0.810$\pm$0.019 & 0.738$\pm$0.018 & 0.847$\pm$0.010\\ 
Backdoor   &0.700$\pm$0.004 & 0.702$\pm$0.034 & 0.678$\pm$0.057 & 0.389$\pm$0.007 & 0.371$\pm$0.025 & \textbf{0.713}$\pm$0.004 & \textbf{0.835}$\pm$0.006 & 0.795$\pm$0.061 & 0.753$\pm$0.049 & 0.804$\pm$0.015 & 0.736$\pm$0.021 & 0.807$\pm$0.009 \\ 
DoS    &0.681$\pm$0.002 & 0.718$\pm$0.022 & 0.690$\pm$0.028 & 0.365$\pm$0.011 & 0.379$\pm$0.027& \textbf{0.751}$\pm$0.011 & 0.809$\pm$0.005 & 0.846$\pm$0.024 & 0.779$\pm$0.030 & 0.757$\pm$0.018 & 0.737$\pm$0.024 & \textbf{0.871}$\pm$0.011 \\ 
Exploits   & \textbf{0.768}$\pm$0.004 & 0.660$\pm$0.046 & 0.636$\pm$0.039 & 0.373$\pm$0.007 & 0.367$\pm$0.026 & 0.589$\pm$0.007 & \textbf{0.906}$\pm$0.004 & 0.871$\pm$0.028 & 0.798$\pm$0.029 & 0.774$\pm$0.021 & 0.732$\pm$0.022 & 0.858$\pm$0.012\\ 
Fuzzers    &\textbf{0.646}$\pm$0.008 & 0.493$\pm$0.028 & 0.499$\pm$0.058 & 0.361$\pm$0.006 & 0.371$\pm$0.025 & 0.500$\pm$0.054 & \textbf{0.878}$\pm$0.002 & 0.841$\pm$0.004 & 0.839$\pm$0.010 & 0.767$\pm$0.015 & 0.735$\pm$0.019 & 0.856$\pm$0.015\\ 
Generic    &\textbf{0.759}$\pm$0.004 & 0.748$\pm$0.039 & 0.703$\pm$0.013 & 0.485$\pm$0.007 & 0.380$\pm$0.026& 0.647$\pm$0.021 & 0.827$\pm$0.007 & 0.820$\pm$0.032 & 0.793$\pm$0.026 & \textbf{0.854}$\pm$0.007 & 0.737$\pm$0.020 & 0.828$\pm$0.014\\
Reconnaissance   & 0.438$\pm$0.009 & 0.386$\pm$0.004 & 0.392$\pm$0.005 & \textbf{0.457}$\pm$0.008 & 0.374$\pm$0.024 &0.414$\pm$0.014 & 0.809$\pm$0.008 & 0.819$\pm$0.002 & 0.821$\pm$0.003 & 0.809$\pm$0.014 & 0.735$\pm$0.012 &\textbf{0.829}$\pm$0.005 \\\hline
Hypothyroid &  \textbf{0.490}$\pm$0.001 & 0.469$\pm$0.005 & 0.398$\pm$0.012 & 0.081$\pm$0.003 & 0.155$\pm$0.020 & 0.432$\pm$0.011 & \textbf{0.846}$\pm$0.001 & 0.835$\pm$0.003 & 0.809$\pm$0.005 & 0.536$\pm$0.009 & 0.683$\pm$0.023 & 0.828$\pm$0.005\\ 
Subnormal &   \textbf{0.436}$\pm$0.007 & 0.379$\pm$0.031 & 0.308$\pm$0.035 & 0.079$\pm$0.002 & 0.184$\pm$0.028 & 0.288$\pm$0.032 & \textbf{0.821}$\pm$0.001 & 0.784$\pm$0.008 & 0.758$\pm$0.005 & 0.523$\pm$0.009 & 0.733$\pm$0.023 & 0.800$\pm$0.009\\ \hline
Downstairs &  \textbf{0.943}$\pm$0.001 & 0.874$\pm$0.025 & 0.887$\pm$0.018 & 0.300$\pm$0.005 & 0.368$\pm$0.016 & 0.844$\pm$0.022 & \textbf{0.993}$\pm$0.001 & 0.990$\pm$0.004 & 0.991$\pm$0.004 & 0.911$\pm$0.006 & 0.926$\pm$0.005 & 0.991$\pm$0.002\\ 
Upstairs &   \textbf{0.942}$\pm$0.005 & 0.865$\pm$0.009 & 0.887$\pm$0.008 & 0.297$\pm$0.004 & 0.394$\pm$0.019 & 0.900$\pm$0.011 & \textbf{0.996}$\pm$0.001 & 0.983$\pm$0.009 & 0.990$\pm$0.001 & 0.918$\pm$0.006 & 0.940$\pm$0.004& 0.994$\pm$0.001 \\ \hline
Cottonwood &   \textbf{0.709}$\pm$0.001 & 0.670$\pm$0.022 & 0.678$\pm$0.028 & 0.424$\pm$0.041 & 0.443$\pm$0.069 & 0.593$\pm$0.023 & \textbf{0.923}$\pm$0.002 & 0.868$\pm$0.019 & 0.876$\pm$0.042 & 0.891$\pm$0.019 & 0.849$\pm$0.027 & 0.841$\pm$0.032\\ 
Douglas-fir    &\textbf{0.776}$\pm$0.003 & 0.722$\pm$0.019 & 0.728$\pm$0.014 & 0.453$\pm$0.021 & 0.456$\pm$0.075& 0.669$\pm$0.011 & \textbf{0.976}$\pm$0.000 & 0.974$\pm$0.003 & 0.973$\pm$0.002 & 0.901$\pm$0.010 & 0.862$\pm$0.028 & 0.963$\pm$0.002 \\\hline
\textbf{Average}  & \textbf{0.690}$\pm$0.004 & 0.641$\pm$0.024 & 0.621$\pm$0.027 & 0.347$\pm$0.010 & 0.355$\pm$0.031 & 0.613$\pm$0.023 & \textbf{0.882}$\pm$0.003 & 0.867$\pm$0.019 & 0.842$\pm$0.017 & 0.789$\pm$0.013 & 0.780$\pm$0.019 & 0.871$\pm$0.010\\ 
\textbf{P-value} & \centering  - & \centering  0.0024 & \centering 0.0005& \centering 0.0005& \centering 0.0002& \centering 0.0034 &\centering -& \centering 0.0254 &  \centering 0.0017 & \centering 0.0010& \centering 0.0002&   0.0769\\\hline
\end{tabular}
}
\label{tab:aucprroc}
\end{table*}

\begin{figure*}[h!]
  \centering
    \includegraphics[width=1.0\textwidth]{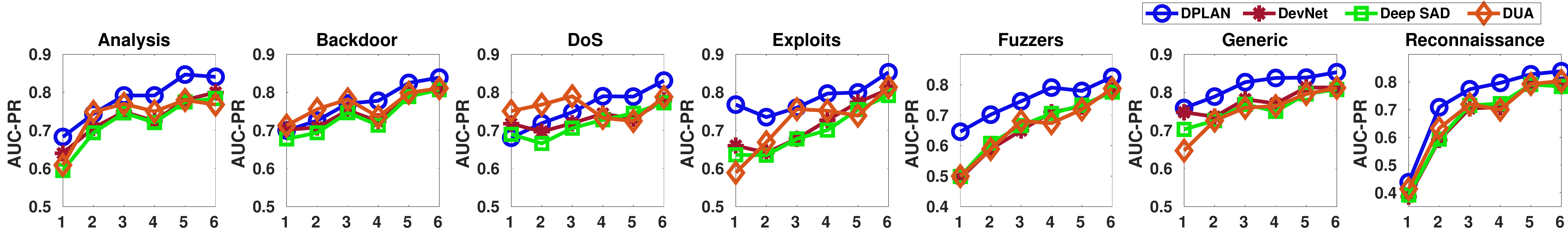}
  \caption{AUC-PR results w.r.t. the number of known anomaly classes. This experiment is inapplicable to the other six datasets.}
  \label{fig:moreanomalyclasses}
\end{figure*}


\subsubsection{Scenario II: Increasing the Number of Known Anomaly Classes}\label{subsec:increasingknownclasses}
We further examine the scenarios with more known anomaly classes. This experiment focuses on the seven \textit{NB15} datasets, since it is inapplicable to \textit{Thyroid}, \textit{HAR} and \textit{Covertype} that contain two anomaly classes only. Particularly, each of these seven datasets is used as a base, and a new randomly selected anomaly class with 60 anomalies is incrementally added into $\mathcal{D}^a$ each step. This results in additional 35 datasets where each training data contains two-to-six known anomaly classes. The test data remains unchanged with seven anomaly classes. \textit{The number of unknown anomaly classes in each data decreases with increasing number of known anomaly classes.}

The AUC-PR results are shown in Figure \ref{fig:moreanomalyclasses}. In general, increasing the coverage of known anomalies provide more supervision information, which enables DPLAN, DevNet, Deep SAD and DUA to achieve considerable improvement, especially on datasets, \eg, \textit{Fuzzers} and \textit{Reconnaissance}, where the first known anomaly class cannot provide much generalizable information. The AUC-PR of DPLAN increases remarkably from 0.438-0.768 up to 0.826-0.853 across the datasets, with maximal relative improvement as large as 91\%. Although DPLAN is less effective than, or entangled with, DevNet, Deep SAD and DUA at the starting point on some datasets, \eg, \textit{Backdoor}, \textit{DoS} and \textit{Generic}, it is improved quickly and finally achieves about 4\%-8\% consistent improvement on those data. 


\subsubsection{Tolerance to Increasing Anomaly Pollution}\label{subsec:contamination}

This section examines the effect of increasing the anomaly contamination/pollution rate. The 13 datasets in Table \ref{tab:aucprroc} serve as our bases. We then incrementally add more unlabeled anomalies into the training data with an anomaly pollution factor of $n\times 2\%$ for each dataset, with $n\in \{1, 2, 3, 4, 5\}$. Combining with the original 2\% pollution, we evaluate six anomaly pollution factors $n\in \{1, 2, 3, 4, 5, 6\}$, resulting in an anomaly contamination rate ranging from 2\% to 12\%. 

\begin{figure*}[h!]
  \centering
    \includegraphics[width=1.0\textwidth]{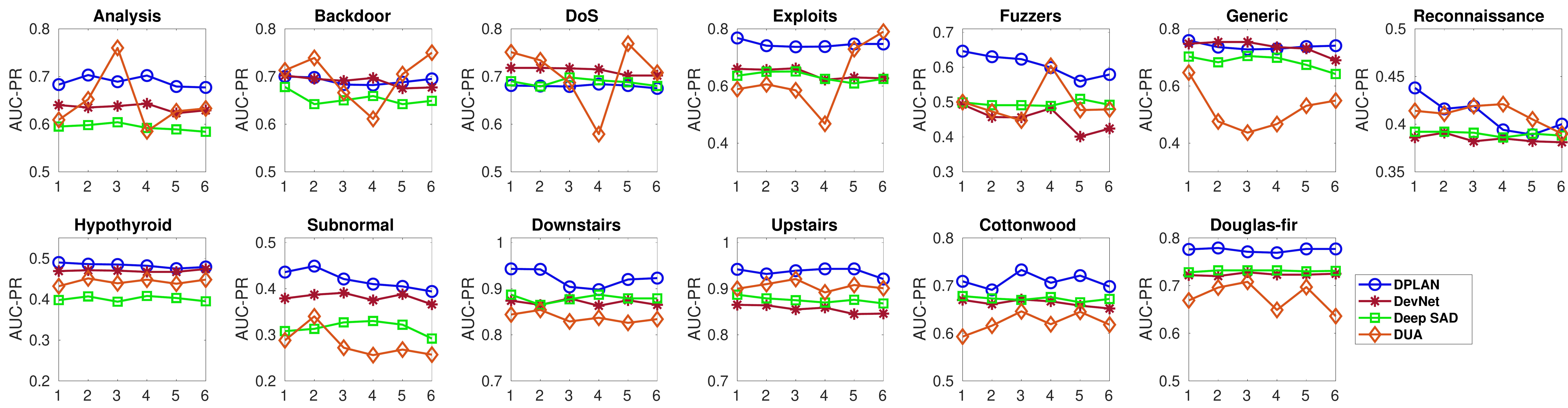}
  \caption{AUC-PR performance w.r.t. different anomaly pollution factors. }
  \label{fig:morenoises}
\end{figure*}

The obtained AUC-PR results are reported in Figure \ref{fig:morenoises}. The following three remarks can be made from the results. (i) Despite of different anomaly pollution, the superiority of DPLAN over the three competing methods is consistent to the results in Table \ref{tab:aucprroc}. (ii) It is interesting that DPLAN, DevNet and Deep SAD perform stably with increasing anomaly pollution factors on several datasets, \ie, \textit{Analysis}, \textit{DoS}, \textit{Hypothyroid}, \textit{Downstairs} and \textit{Douglas-fir}, while having clear downward trends on the rest of the other datasets where the supervisory signal from the labeled anomalies may not be strong enough to tolerate the noises. (iii) DUA demonstrates largely fluctuated performance. This may be due to the unstable performance of its pseudo labeling module. 

\subsubsection{Comparison Summary}

We summarize and discuss the empirical results in this section from three aspects as follows.
\begin{itemize}
    \item DPLAN vs. DevNet and Deep SAD. Compared to these two semi-supervised detectors that exclusively learn the known abnormality, DPLAN is driven to learn more effective known abnormality through its unique automatic agent-environment interactions; moreover, it is able to learn the abnormality lying beyond the span set of the given anomaly examples, for which DevNet and Deep SAD fail to do so. These two unique advantages enable DPLAN to gain significantly better accuracy (\eg, 7\%-11\% in AUC-PR) and more stable performance in diverse scenarios, as shown in Table \ref{tab:aucprroc}, Figures \ref{fig:moreanomalyclasses} and \ref{fig:morenoises}.
    \item DPLAN vs. REPEN and iForest. DPLAN is anomaly-informed and is able to achieve nearly 100\% improvement over two state-of-the-art unsupervised detectors. As shown in Table \ref{tab:aucprroc}, DevNet and Deep SAD can also achieve large improvement over REPEN and iForest, which is consistent to \cite{pang2019devnet,ruff2020deep}. The unsupervised detectors may perform better than the anomaly-informed detectors in some cases where their underlying intuition of anomalies is well matched to that in the datasets, \eg, REPEN on \textit{Generic} and \textit{Reconnaissance}.
    \item DPLAN vs. DUA. As shown in both Table \ref{tab:aucprroc}, Figures \ref{fig:moreanomalyclasses} and \ref{fig:morenoises}, the unsupervised anomaly labeling in DUA helps improve DevNet in a few datasets such as \textit{Backdoor} and \textit{DoS}.
    However, it leads to significantly degraded and fluctuated DevNet in most datasets, especially in AUC-PR.
    Although DPLAN and DUA use the same unsupervised detector to explore the unlabeled data, DPLAN learns to automatically balance the exploitation of the anomaly examples and the exploration of the unlabeled data, jointly optimizing known and unknown anomaly detection. This results in substantially better performance than the decoupled two-step approach in DUA.
\end{itemize}



\subsection{Empirical Analysis of DPLAN}

\subsubsection{Ablation Study}\label{subsec:ablation}

To understand the contribution of the key modules of DPLAN, it is compared with its three ablated variants:
\begin{itemize}
    \item \textbf{ERew}. ERew is DPLAN with the External Reward (ERew) only, \ie, the intrinsic reward function $f$ is removed.
    \item \textbf{REnv}. REnv is ERew with the anomaly-biased observation sampling function $g$ replaced with a Random Environment (REnv), \ie, the observation is randomly sampled from $\mathcal{D}$.
    \item \textbf{DQN}$^+$. DQN$^+$ is DPLAN with a deeper Q-network. Two additional hidden layers with respective 500 and 100 ReLU units are added, with each followed by a dropout layer with the same drop rate of 0.9.
\end{itemize}

\begin{table}[htbp]
  \centering
  \caption{Results of DPLAN (Org) and its ablated variants}
  \scalebox{0.75}{
    \begin{tabular}{  l@{}cccc|cccc  }
    \hline
          & \multicolumn{4}{c|}{\textbf{AUC-PR Performance}} & \multicolumn{4}{c }{\textbf{AUC-ROC Performance}} \\\hline
    \textbf{Dataset}  & \textbf{Org} & \textbf{ERew} & \textbf{REnv} & \textbf{DQN}$^+$ & \textbf{Org} & \textbf{ERew} & \textbf{REnv} & \textbf{DQN}$^+$ \\\hline
    Analysis & 0.683 & 0.661 & 0.674 & \textbf{0.776} & 0.852 & 0.854 & 0.795 & \textbf{0.911} \\
    Backdoor & 0.700 & 0.680 & 0.677 & \textbf{0.820} & 0.835 & 0.769 & 0.776 & \textbf{0.921} \\
    DoS & 0.681 & 0.676 & 0.696 & \textbf{0.758} & 0.809 & 0.779 & 0.806 & \textbf{0.917} \\
    Exploits & \textbf{0.768} & 0.751 & 0.679 & 0.720 & \textbf{0.906} & 0.903 & 0.785 & 0.898 \\
    Fuzzers & 0.646 & 0.637 & \textbf{0.697} & 0.686 & 0.878 & 0.867 & 0.808 & \textbf{0.887} \\
    Generic & 0.759 & \textbf{0.779} & 0.643 & 0.729 & 0.827 & \textbf{0.869} & 0.737 & 0.832 \\
    Reconnaissance & 0.438 & 0.446 & \textbf{0.665} & 0.540 & 0.809 & 0.787 & 0.766 & \textbf{0.879} \\\hline
    Hypothyroid & \textbf{0.490} & 0.485 & 0.119 & 0.443 & \textbf{0.846} & 0.844 & 0.633 & 0.816 \\
    Subnormal & 0.436 & \textbf{0.447} & 0.118 & 0.245 & \textbf{0.821} & 0.818 & 0.624 & 0.758 \\\hline
    Downstairs & \textbf{0.943} & 0.941 & 0.268 & 0.869 & \textbf{0.993} & 0.981 & 0.800 & 0.989 \\
    Upstairs & \textbf{0.942} & 0.927 & 0.193 & 0.881 & \textbf{0.996} & 0.983 & 0.695 & 0.991 \\\hline
    Cottonwood & 0.709 & \textbf{0.722} & 0.367 & 0.700 & 0.923 & 0.939 & 0.807 & \textbf{0.937} \\
    Douglas-fir & \textbf{0.776} & 0.765 & 0.300 & 0.752 & \textbf{0.976} & 0.975 & 0.747 & 0.975 \\\hline
    \textbf{Average} & \textbf{0.690} & 0.686 & 0.469 & 0.686 & 0.882 & 0.874 & 0.752 & \textbf{0.901} \\\hline
    \textbf{P-value} & - &  0.4490&  0.0215 &  1.0000   & - & 0.0903 & 0.0002 & 0.2812 \\ \hline
    \end{tabular}%
    }
  \label{tab:ablation}%
\end{table}%

The comparison results are provided in Table \ref{tab:ablation}. Despite two losses on \textit{Generic} and \textit{Cottonwood}, Org outperforms ERew on most datasets, especially in AUC-ROC for which Org is significantly better than ERew at the 90\% confidence level. This indicates that the intrinsic reward module enables DPLAN to well balance the exploration of the unlabeled data and the exploitation of the labeled anomalies. On the other hand, ERew also significantly outperforms DevNet and Deep SAD in Table \ref{tab:aucprroc}. This demonstrates that DPLAN (with the external reward alone) can learn the known abnormality significantly better than DevNet and Deep SAD.

Compared to REnv, Org gains more than 47\% and 17\% average AUC-PR and AUC-ROC improvement, respectively, demonstrating the great importance of the anomaly-biased environment. Impressively, DQN$^+$ achieves remarkably improvement over Org. This is very encouraging because it indicates DPLAN can learn more complex yet well generalized models from the limited labeled data when the amount of unlabeled data is large, \eg, the seven \textit{UNSW\_NB15} datasets and the two \textit{Covetype} datasets, while prior methods like DevNet drop significantly when using a deeper architecture \cite{pang2019devnet},

\subsubsection{Learning with More Training Steps}

\begin{figure}[h!]
  \centering
    \includegraphics[width=0.485\textwidth]{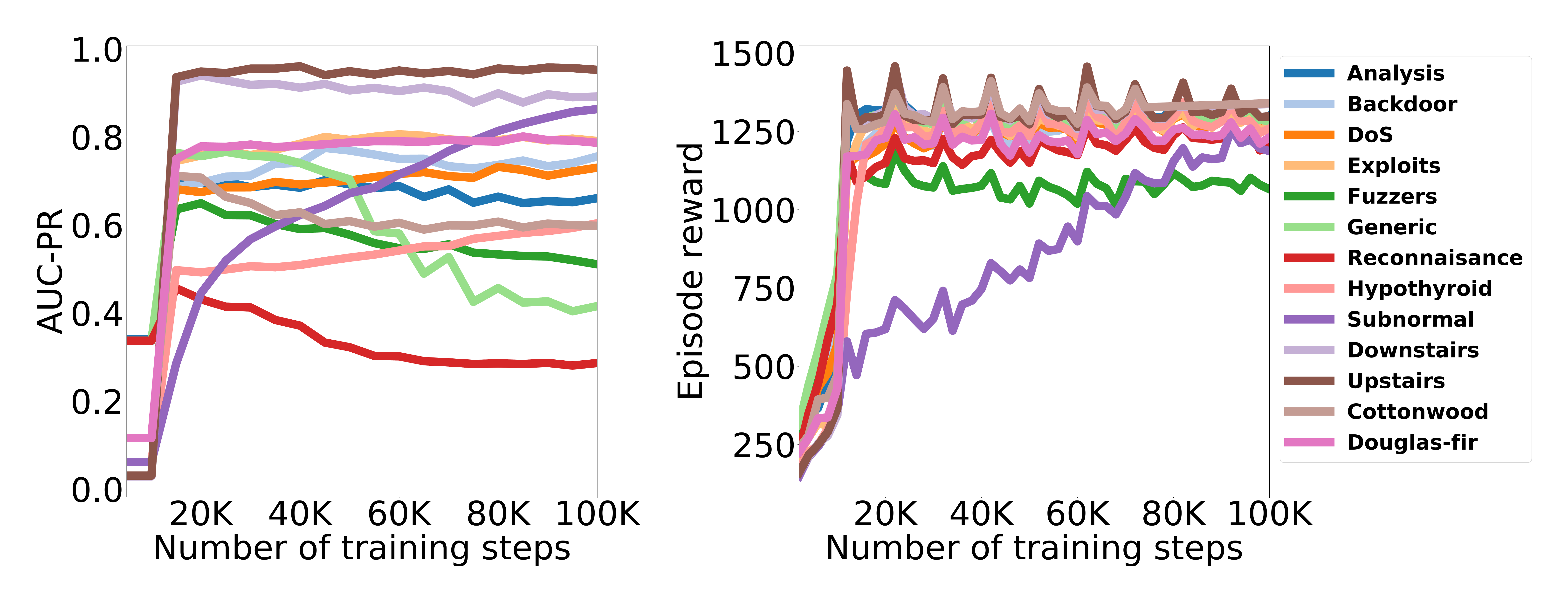}
  \caption{AUC-PR and episode reward w.r.t. training steps}
  \label{fig:moresteps}
\end{figure}

We investigate the capability of DPLAN in further lifting the performance with increasing reinforcement steps. The results are given in Figure \ref{fig:moresteps}. It shows that the AUC-PR and episode reward of DPLAN often converge very early, \eg, around 20,000 training steps, resulting in stable superior performance across all 13 datasets at that point. It is very interesting that through the 100,000 training steps, DPLAN is continuously enhanced on the \textit{Hypothyroid} and \textit{Subnormal} datasets, increasing the AUC-PR from 0.490 and 0.436 up to 0.604 and 0.863 respectively. This results in as large as further 23\% and 98\% AUC-PR improvement compared to the version trained with 20,000 steps. This indicates that with larger training steps, DPLAN may achieve better exploration on these two datasets. However, the opposite may occur on the three other datasets \textit{Fuzzers}, \textit{Generic} and \textit{Reconnaissance}. DPLAN trained with 20,000 steps is generally recommended. 

\subsubsection{Sensitivity w.r.t. Representation Dimensionality Size}\label{subsubsec:sensitivity}

\begin{figure}[htbp]
 		\centering
    \includegraphics[width=0.50\textwidth]{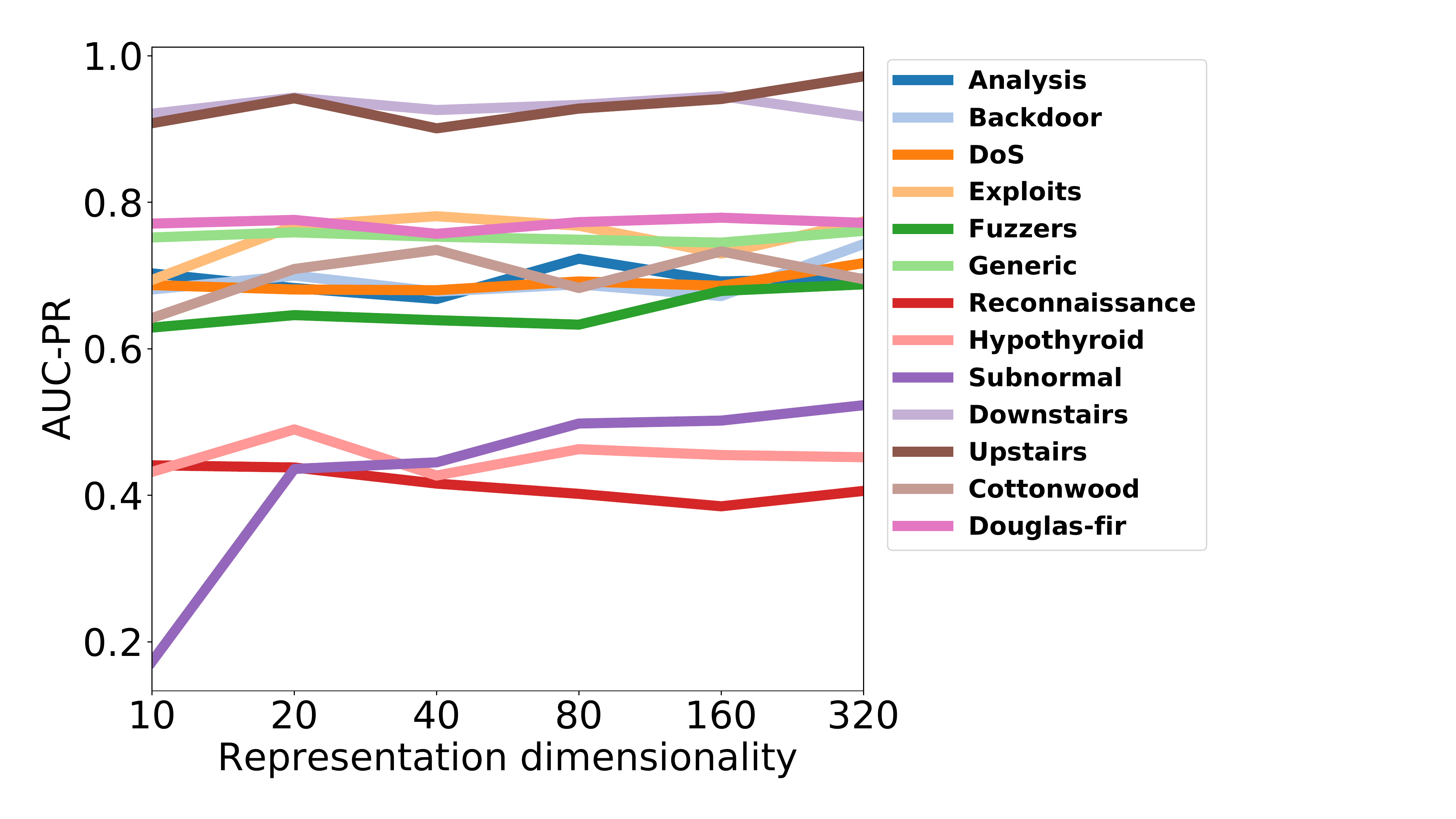}
  \captionof{figure}{AUC-PR w.r.t. representation dimensionality}\label{fig:embedding_size}
\end{figure}

We also examine the sensitivity of DPLAN w.r.t. the representation dimensionality size in its feature layer. A set of dimensionality sizes in a large range, \ie, $\{10, 20, 40, 80, 160, 320\}$, is used. The AUC-PR results on all the 13 datasets are shown in Figure \ref{fig:embedding_size}. In general, DPLAN performs rather stably with different dimensionality sizes across the datasets. DPLAN performs less effectively using only 10 representation dimensions. DPLAN performs stably using 20 representation dimensions; increasing the dimensionality size does not change the performance much. This may be due to that the supervisory information that can be leveraged by DPLAN is bounded at some point. In some cases where more supervisory information can be leveraged for building more complex models, such as on \textit{Subnormal} (as observed in Figure \ref{fig:moresteps}), the performance of DPLAN is continuously improved with increasing dimensionality. 

\subsubsection{Computational Efficiency}\label{subsubsec:efficiency} The runtime of training DPLAN is constant w.r.t. data size due to the stochastic optimization used and is linear to the number of training steps. It takes 15 to 120 seconds on most of the datasets in Table \ref{tab:aucprroc} to train the anomaly detection agent in the default DPLAN. This is often slower than the competing methods since DPLAN has a large number of interactions with the environment. Fortunately, in practice the model training can be easily taken offline. The online detection runtime is normally more important. Similar to DevNet and Deep SAD, DPLAN takes a single forward-pass to obtain the anomaly scores, so these three methods have the same online time complexity. They takes less than three seconds to complete the anomaly scoring of over 27,5000 test instances in total in all 13 datasets, which is faster than REPEN and iForest that respectively takes about 20 and 40 seconds.

\section{Conclusions}

This paper proposes an anomaly detection-oriented deep reinforcement learning framework and its instantiation DPLAN. Our anomaly detection-oriented agent is driven by an labeled anomaly data-based external reward to learn the known abnormality while at the same time actively exploring unknown anomalies in unlabeled data to continuously refine the learned abnormality. This enables DPLAN to learn significantly more generalized abnormality than existing methods. The better generalizability also allows us to build more effective DPLAN with a deeper network architecture, which is not viable to the competing methods. Impressively, DPLAN can achieve further 23\%-98\% AUC-PR improvement over its default version by only increasing the number of training steps on some datasets. Its inference is also computationally efficient to scale. 

\bibliographystyle{ACM-Reference-Format}
\balance
\bibliography{references}

\newpage
\appendix
\section{Supplementary Material for Reproducibility}
\subsection{Data Accessing and Preprocessing}\label{sec:preprocess}

UNSW\_NB15 is a recently released network intrusion datasets with a set of network attacks. The seven most common types of attacks, including \textit{analysis}, \textit{backdoor}, \textit{DoS}, \textit{exploits}, \textit{fuzzers}, \textit{generic} and \textit{reconnaissance}, are the anomalies against the normal network flows. Annthyroid is a dataset for detection of the thyroid diseases, in which patients diagnosed with \textit{hypothyroid} or \textit{subnormal} are anomalies. HAR contains embedded inertial sensor data from a waist-mounted smartphone for six different human activities. The activities of walking downstairs and walking upstairs (\textit{downstairs} and \textit{upstairs} for short) are treated as abnormal activities w.r.t. the other four common activities. Covertype contains cartographic data of seven forest cover types. Following the literature \cite{liu2012iforest,ting2017defying,siddiqui2018kdd,akoglu2012fast}, the most dominant cover type \textit{lodgepole pine} is used as the normal class against \textit{cottonwood} and \textit{douglas-fir} that demonstrates obvious deviations from lodgepole pine. Following the literature \cite{ting2017defying,liu2012iforest,pang2019devnet,akoglu2012fast}, random downsampling without replacement is applied to the \textit{DoS}, \textit{exploits}, \textit{fuzzers}, \textit{generic}, \textit{reconnaissance}, \textit{downstairs} and \textit{upstairs} classes to guarantee the rarity nature of anomalies. Specifically, we downsample the \textit{DoS}, \textit{exploits}, \textit{fuzzers}, \textit{generic}, \textit{reconnaissance} anomaly classes to have 3,000 instances so that all anomaly classes in the UNSW\_NB15 data are of a similar size. This is to guarantee the class balance among the anomaly classes to have fair evaluation of the performance in detecting anomalies from different anomaly classes. The \textit{downstairs} and \textit{upstairs} are downsampled so that the anomalies from each of these classes account for 2\% of the dataset.

One-hot encoding is used to convert all categorical features into numeric features. Missing values are replaced with the mean value if there are any features containing missing values. All features are normalized into the range $[0,1]$ before modeling. These four datasets are publicly available and can be accessed via the links given in Table \ref{tab:link}.

\begin{table}[htbp]
\caption{Links for Accessing the Data Sets}
\scalebox{0.78}{
    \begin{tabular}{l|p{8.0cm}}
    \hline
    \textbf{Data} & \textbf{Link}\\\hline
    UNSW\_NB15&https://www.unsw.adfa.edu.au/unsw-canberra-cyber/cybersecurity/ADFA-NB15-Datasets/ \\ 
    Annthyroid&https://www.openml.org/d/40497 \\
    HAR&https://www.openml.org/d/1478 \\ 
    Covertype&https://archive.ics.uci.edu/ml/datasets/covertype \\ \hline
    \end{tabular}
}
\label{tab:link}
\end{table}

\subsection{The Algorithm of DPLAN}

The procedure of training DPLAN is presented in Algorithm \ref{alg:deepquad}. The first three steps initialize the size of the experience set and weight parameters of Q-value functions $Q$ and $\hat{Q}$. DPLAN is then trained with $\mathit{n\_episodes}$ episodes, with each episode $\mathit{n\_steps}$ training steps. For each episode, the first observation $\mathbf{s}_1 \sim U(\mathcal{D}^{u})$ is uniformly sampled at random from the unlabeled data $\mathcal{D}^{u}$. In Steps 7-8, we adopt the same $\epsilon$-greedy exploration as in the original DQN, in which with a probability of $\epsilon$ the agent randomly selects an action from $\{a^0,a^1\}$, and otherwise selects the action that maximizes the action-value function at the current time step. After the agent performing the selected action, the environment responses to the agent with next observation $\mathbf{s}_{t+1}$, with probability $p$ we randomly sample it from the labeled anomaly set $\mathcal{D}^{a}$, \ie, $\mathbf{s}_{t+1} \sim U(\mathcal{D}^{a})$, and otherwise return the nearest/farthest neighbor of $\mathbf{s}_t$ in a random subsample $\mathcal{S}\subset\mathcal{D}^{u}$ based on $g_{u}(\mathbf{s}_{t+1}|\mathbf{s}_t, a_t;\theta^e)$, where $\theta^e$ is a subset of parameters in $\theta$ and is constantly updated. The environment also gives a reward $r_t^e$ to the agent, with $r_t^e$ calculated by Eqn. (\ref{eqn:externalreward}). At the same time, $f(\mathbf{s}_t,\hat{\theta}^{e})$ is used to yield an intrinsic reward $r^i_t$. $\hat{\theta}^{e}$ is exactly the same set of parameters as $\theta^e$, but we update $\hat{\theta}^{e}=\theta^e$ every $N$ steps rather than every step. Constantly updating $\hat{\theta}^{e}=\theta^e$ requires to frequently project data onto low-dimensional space and build iForest, which adds remarkably extra computation. We then combine the two rewards by $r_t = r^e_t + r^i_t$ in Step 12, \ie, the agent always receives a combined reward for each observation $\mathbf{s}_t$, regardless of $\mathbf{s}_t\in \mathcal{D}^{a}$ or $\mathbf{s}_t\in \mathcal{D}^{u}$. After that, we gain an experience record $(\mathbf{s}_t,a_t,r_t,\mathbf{s}_{t+1})$ and store it into the experience set $\mathcal{E}$. Steps 14-16 then performs the Q-learning update, with the target action-value function $\hat{Q}=Q$ reset every $K$ steps.

\renewcommand{\algorithmicrequire}{\textbf{Input:}}
\renewcommand{\algorithmicensure}{\textbf{Output:}}
\begin{algorithm}
\small 
\caption{\textit{Training DPLAN}}
\begin{algorithmic}[1]
\label{alg:deepquad}
\REQUIRE $\mathcal{D}=\{\mathcal{D}^{a},\mathcal{D}^{u}\}$ - training data
\ENSURE $Q(\mathbf{s},a;\theta^{*})$ - action-value function (anomaly detection agent)
\STATE Initialize action-value function $Q$ with random weights $\theta$
\STATE Initialize target action-value function $\hat{Q}$ with weights $\theta^{-}=0$
\STATE Initialize the size of experience set $\mathcal{E}$ to $M$
\FOR{ $j = 1$ to $\mathit{n\_episodes}$}
    \STATE Initial observation $\mathbf{s}_1 \sim U(\mathcal{D}^{u})$
    \FOR{ $t = 1$ to $\mathit{n\_steps}$}
        \STATE With probability $\epsilon$ select a random action $a_t$ from $\{a^0,a^1\}$
        \STATE Otherwise select $a_t = \argmax_{a}Q(\mathbf{s}_t,a;\theta)$ 
        \STATE With probability $p$ the environment returns $\mathbf{s}_{t+1} \sim U(\mathcal{D}^{a})$
        \STATE Otherwise return $\mathbf{s}_{t+1} \sim \mathcal{D}^{u}$ based on $g_{u}(\mathbf{s}_{t+1}|\mathbf{s}_t, a_t;\theta^e)$
        \STATE Calculate intrinsic reward $r^i_t = f(\mathbf{s}_t,\hat{\theta}^{e})$ 
        \STATE Receive reward $r_t = r^e_t + r^i_t$
        \STATE Store experience $(\mathbf{s}_t,a_t,r_t,\mathbf{s}_{t+1})$ in $\mathcal{E}$
        \STATE Randomly sample minibatch of experience records $(\mathbf{s}_l,a_l,r_l,\mathbf{s}_{l+1})$ from $\mathcal{E}$
        \STATE \[ \text{Set} \; y_l =
        \begin{cases}
             r_l    \quad \quad \text{if episode terminates at step } l+1 \\
             r_l + \gamma \max_{a^\prime} \hat{Q} (\mathbf{s}_{l+1},a^\prime;\theta^{-}) \quad \text{otherwise }
        \end{cases}
        \]
        \STATE Perform a gradient descent step on $\big(y_l - Q (\mathbf{s}_{l},a_{l};\theta)\big)^2$ w.r.t. the weight parameters $\theta$
        \STATE Update $\hat{\theta}^{e}= \theta^e $ every $N$ steps
        \STATE Update $\hat{Q}=Q$ every $K$ steps
    \ENDFOR
\ENDFOR
\RETURN $Q$
\end{algorithmic}
\end{algorithm}

After training, DPLAN returns $Q(\mathbf{s},a;\theta^{*})$, which is an approximated optimal action-value function and can be seen as an anomaly detection agent to detect anomalies. The procedure of using DPLAN to detect anomalies in a test set $\mathcal{T}$ is presented in Algorithm \ref{alg:evaluation}. Specifically, given every observation $\mathbf{s}_j \in \mathcal{T}$, DPLAN performs one forward-pass in its network and then gets the estimated action-value for each action. If $\mathbf{s}_j$ is believed to be an anomaly, DPLAN would select action $a^1$ with a large action-value, and select $a^0$ with a small action-value otherwise. Thus, the action-value is used as an end-to-end learnable anomaly score measure.

\renewcommand{\algorithmicrequire}{\textbf{Input:}}
\renewcommand{\algorithmicensure}{\textbf{Output:}}
\begin{algorithm}
\small 
\caption{\textit{Anomaly Detection using DPLAN}}
\begin{algorithmic}[1]
\label{alg:evaluation}
\REQUIRE $\mathcal{T}$ - test data, $Q(\mathbf{s},a;\theta^{*})$ - anomaly detection agent
\ENSURE $\mathbf{y}$ - anomaly scores
\FOR{ $j = 1$ to $|\mathcal{T}|$}
    \STATE $y_j = \; \argmax_{a} Q(\mathbf{s}_j,a;\theta^{*})$, $\mathbf{s}_j \in \mathcal{T}$
\ENDFOR
\RETURN Anomaly scores $\mathbf{y}$
\end{algorithmic}
\end{algorithm}

\subsection{Implementation Details}\label{subsec:implementation}

\subsubsection{Algorithm Implementation}

All methods are implemented using Python, with DevNet and REPEN directly taken from the authors at https://sites.google.com/site/gspangsite/sourcecode and iForest taken from the scikit-learn package. Deep anomaly detection methods DevNet, DUA, REPEN and Deep SAD are built upon Keras with Tensorflow as the backend. Oversampling is used in all these four methods to guarantee that we have the same proportion of labeled (or pseudo) anomalies and (pseudo) normal instances in each mini-batch. This is a common method used to tackle the class imbalance issue.

We implement DPLAN based on the deep Q-network implementation in the open-source Keras-based deep reinforcement learning project, namely, Keras-rl, available at https://github.com/keras-rl/keras-rl. Our anomaly-biased simulation environment is implemented under the OpenAI Gym environment. The main packages and their versions used in this work are provided as follows:
\begin{itemize}
    \item gym==0.12.5
    \item keras==2.3.1
    \item keras-rl==0.4.2
    \item numpy==1.16.2
    \item pandas==0.23.4
    \item scikit-learn==0.20.0
    \item scipy==1.1.0
    \item tensorboard==1.14.0
    \item tensorflow==1.14.0
\end{itemize}

All of the runtime results in Section \ref{subsubsec:efficiency} are calculated under the same environment: Intel® Core™ i7-8700 CPU @ 3.20GHz × 12, 16GB RAM.

\subsubsection{Hyperparameter Settings}

The network architecture used in DPLAN, DevNet, Deep SAD, DUA, and REPEN contains one hidden layer with 20 ReLU units by default. The DPLAN with a deeper network architecture (\ie, the variant of DPLAN - DQN$^+$) adds two additional hidden layers immediately after the input layer. The first hidden layer contains 500 ReLU units while the second hidden layer contains 100 ReLU. Following each of these two hidden layers, we add a dropout layer to avoid overfitting. The dropout rate is 0.9 for both dropout layers.

Since original deep Q-network is designed for complex control tasks with a large set of possible actions in very high-dimensional space, some of its recommended parameter settings are not applicable to our anomaly detection task with two possible actions. Therefore, in addition to adapt the network architecture, some parameters also need to be accordingly adapted. Specifically, DPLAN is trained with 20,000 steps by default, with 10,000 warm-up steps and the target network updated every 10,000 steps. Each episode contains 2,000 steps. The episode is terminated only when the 2,000 steps are completed. We update the parameters $\theta^e$ in the intrinsic reward function $f$ every episode (\ie, 2,000 steps). Also, as shown in Algorithms \ref{alg:deepquad} and \ref{alg:evaluation}, the $\epsilon$ greedy exploration is only used in our model training, with $\epsilon$ annealed from 1.0 to 0.1 over the course of 10,000 steps; it is not used in our evaluation since we does not need any further exploration during testing. The experience replay memory size is set to 100,000 since our agent can typically converge very early. The other parameters of deep Q-network are set to the default settings as in the original DQN \cite{mnih2015dqn}, with some key hyperparameter settings shown in Table \ref{tab:parameters}.

\begin{table}[htbp]
\centering
\caption{Key default hyperparameters from original DQN}
\scalebox{0.8}{
    \begin{tabular}{lc}
    \hline
    \textbf{Hyperparameter} & \textbf{Value}\\\hline
    minibatch size & 32\\
    discount factor $\gamma$ & 0.99\\
    learning rate & 0.00025\\
    gradient momentum & 0.95\\
    min squared gradient & 0.01\\\hline
    \end{tabular}
}
\label{tab:parameters}
\end{table}
 
\end{document}